\documentclass[letterpaper, 10 pt, conference]{ieeeconf}  

\IEEEoverridecommandlockouts                              

\makeatletter
\let\NAT@parse\undefined
\makeatother

\usepackage{graphics} 
\usepackage{graphicx}
\usepackage{grffile} 
\usepackage{amsmath} 
\usepackage{amssymb}  
\usepackage{color}
\usepackage{bm}
\usepackage{url}

\usepackage[noend]{algpseudocode}
\usepackage{algorithm}
\usepackage{makecell}
\usepackage{float}
\usepackage[square, comma, numbers,sort&compress]{natbib}
\usepackage{multirow}
\let\labelindent\relax
\usepackage{enumitem}
\usepackage[table,dvipsnames]{xcolor}
\usepackage[pagebackref=true,breaklinks=true,colorlinks,bookmarks=false]{hyperref}
\usepackage{times}
\usepackage{textcomp}
\usepackage{booktabs}
\usepackage{threeparttable} 
\usepackage{siunitx}

\hypersetup{
linkcolor=BrickRed
,citecolor=Green
,filecolor=Mulberry
,urlcolor=NavyBlue
,menucolor=BrickRed
,runcolor=Mulberry
,linkbordercolor=BrickRed
,citebordercolor=Green
,filebordercolor=Mulberry
,urlbordercolor=NavyBlue
,menubordercolor=BrickRed
,runbordercolor=Mulberry
}

\title{\LARGE \bf
Projector-Guided Non-Holonomic Mobile 3D Printing
}

\author{\small Xuchu Xu$^{1}$, Ziteng Wang$^{2}$, Chen Feng$^{1\dagger}$

\thanks{$^{1}$ New York University, Brooklyn, NY 11201, USA
	{\tt\small \{xuchuxu, cfeng\}@nyu.edu}}%
\thanks{$^{2}$
	{\tt\small tonywangziteng@163.com}}%
\thanks{$^{\dagger}$ Chen Feng is the corresponding author.}%
	
}

\begin{document}

\maketitle
\thispagestyle{empty}
\pagestyle{empty}

\begin{abstract}

Fused deposition modeling (FDM) using mobile robots instead of the gantry-based 3D printer enables additive manufacturing at a larger scale with higher speed. This introduces challenges including accurate localization, control of the printhead, and design of a stable mobile manipulator with low vibrations and proper degrees of freedom. We proposed and developed a low-cost non-holonomic mobile 3D printing system guided by a projector via learning-based visual servoing. It requires almost no manual calibration of the system parameters. Using a regular top-down projector without any expensive external localization device for pose feedback, this system enabled mobile robots to accurately follow pre-designed millimeter-level printing trajectories with speed control. We evaluate the system in terms of its trajectory accuracy and printing quality compared with original 3D designs. We further demonstrated the potential of this system using two such mobile robots to collaboratively print a 3D object with dimensions of 80 \textit{cm} $\times$ 30 \textit{cm} size, which exceeds the limitation of common desktop FDM 3D printers.

\end{abstract}


\section{Introduction}

Fused deposition modeling (FDM) is a common type of additive manufacturing (AM) method. A conventional FDM 3D printer is typically implemented as a gantry system enabling 3 degrees-of-freedom (DOFs) control of a printhead to emit fused materials, ceramic, or even concrete precisely to designed printing positions layer by layer. Such a gantry-based printer cannot print large-scale objects, with typically size limitations of $350\times350\times300\ mm^3$ for desktop printers and $900\times600\times900\ mm^3$ for industrial ones. A gantry system also makes it difficult for multiple printers to collaborate to achieve a faster printing speed. To remove these limitations, installing the printhead as the end-effector of a mobile manipulator is an appealing option.

However, a mobile 3D printer brings several challenges to the software and hardware design of such a robotic system. First, we can no longer benefit from stepper motors on the gantry to estimate the printhead position, which is needed for the computer numerical control of the printer. For a desktop FDM printer with a millimeter-level nozzle diameter, the localization and control of the printhead must achieve a millimeter or even sub-millimeter level to ensure successful printing. Otherwise, the layer-by-layer mechanism could easily fail before completing the printing.

\begin{figure}[!t]
    \centering
    \includegraphics[width=1\columnwidth]{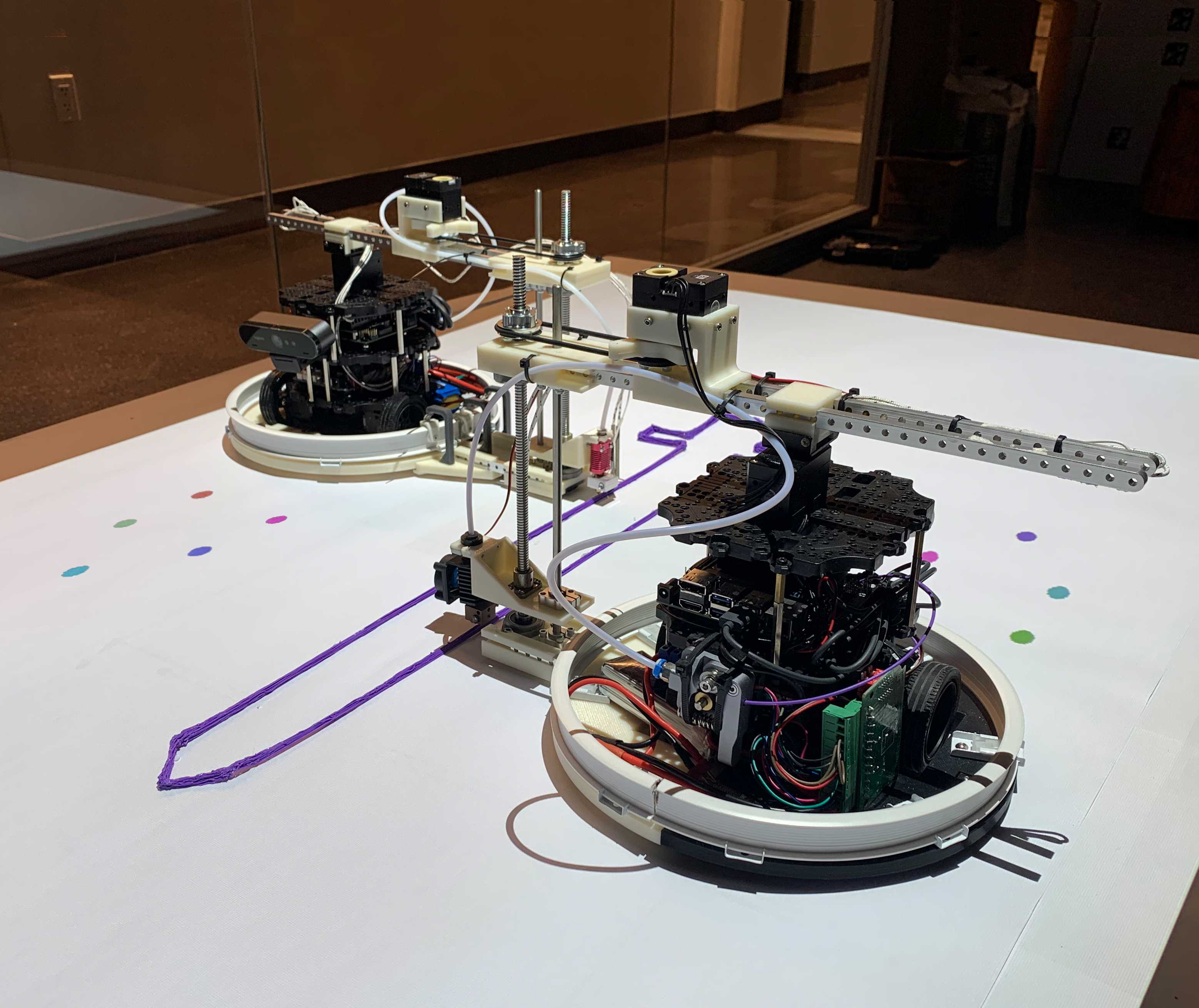}
    \caption{Collaborative printing of two mobile 3D printers in our system.}
    \label{Eye-catchin}
    \vspace{-5mm}
\end{figure}

Although millimeter-level positioning in a room-scale indoor environment can be achieved by camera-based precise localization systems, such as Vicon\textsuperscript{\textregistered}, the issues of such a solution are the system cost and the non-trivial setup and calibration. 
To support easy and friendly deployment, is it possible to develop a cost-efficient localization and control system for our mobile 3D printer without a complex manual system calibration process?

In addition to the software challenge, the hardware design of the robotic system is also non-trivial. Unlike regular mobile manipulators, which are often used for object gripping tasks, the aforementioned accuracy requirement means that the mobile platform must have good stability during its movement to minimize the vibration of the printhead. This could lead to a trade-off with the convenience of the mobile platform's planning and control: should we use a holonomic or a non-holonomic mobile robot?

To address these challenges, we propose and develop a low-cost non-holonomic mobile 3D printing system. The localization and control system is guided by a regular low-cost top-down projector equipped with learning-based visual servoing. The mobile platform is a non-holonomic two-wheel differential drive robot with a camera for visual servoing.

The following are our main contributions: 
\begin{itemize}
	\item We propose a low-cost, projector-guided, non-holonomic mobile 3D printing system with almost no manual calibration and easier ground settings.
	\item We present a learning-based visual servoing method for differential drive robot control with millimeter-level trajectory accuracy.
	\item We conduct both simulated and real-world experiments to validate our system prototype in terms of trajectory control accuracy and printing quality.
	\item We successfully printed a 3D object with dimensions of $800\ mm \times300\ mm$ using two prototype mobile printers.
\end{itemize} 


\section{Related Work}

\textbf{Mobile 3D printing.}
The accuracy of traditional gantry-based 3D printing relies on counting the steps of a stepper motor's output. For mobile 3D printing, the biggest challenge is how to localize a robot, because the accuracy and reliability of the wheel encoder and Inertial measurement unit (IMU) cannot provide satisfactory position feedback. Therefore, some pioneers have explored different types of localization methods for mobile 3D printing systems.

\citet{marques2017mobile} proposed an omni-wheel 3D printing robot and a grid-based 3D printing system. Their mobile robot scans a grid on the ground through an optical sensor to perceive its position when printing. In addition, this system allows multiple robots to perform collaborative printing by operating on the same grid building plane. To avoid collisions in collaborative printing, they presented an automatically generated scheduling approach based on the dependency tree in~\cite{poudel2020generative}. To solve the power supply problem, \citet{currencefloor} invented a floor power module with installed brushes on the robot that connected to the powered floor surface. 

\citet{zhang2018large} used a holonomic-based mobile manipulator for mobile 3D concrete printing. They cut the printing model into several parts and moved the mobile platform to the target workspace for printing execution. They later updated their control algorithm ~\cite{tiryaki2018printing}. This algorithm could obtain localized feedback and adjust the motion error from the AprilTags~\cite{olson2011apriltag} on the ground. Since concrete needs to be pumped through a pipe, \citet{zhang2019planning} designed an algorithm that included four different modes to solve the motion planning problem for multiple tethered planar mobile robots. Similar to~\cite{zhang2018large}, \citet{lv2019large} proposed a holonomic mobile printing method by switching the workspace and used the contour of the printed object to estimate the pose of the mobile platform and update the mobile platform control.

\textbf{Holonomic vs. non-holonomic mobile platform.} The main differences between the above mobile 3D printing systems and ours are twofold. First, we use a non-holonomic robot as our printing platform. According to previous printing results~\cite{marques2017mobile}, although omni-wheel-based holonomic robots can be more convenient in terms of control and path planning, they still cannot avoid shifting between printed layers caused by a \textit{slip on the orthogonal direction of the robot motion}. Our differential drive robot uses standard wheels and does not suffer from slippage in that direction. This also allows odometry based on wheel encoders, which is more difficult in omni-wheel robots. The second difference is the localization system. Their printing platforms operate in an absolute Cartesian coordinate frame. The accuracy of those systems relies on the grid, \textit{which requires manual setup on the ground}. Our system uses projector-based visual servoing control, which requires no robot pose feedback in a Cartesian frame, nor do we need any non-trivial manual calibration or setup.

\textbf{Visual servoing control.}
Prior work on visual servoing (VS) control generally fell into three categories: position-based, image-based, and fusion approaches~\cite{chaumette2006visual,malis19992}. These three types of visual servoing methods control the robot movement by reducing the error between the observed image obtains error feedback from different spaces.

Position-based visual servoing (PBVS)~\cite{weiss1987dynamic,wilson1996relative,thuilot2002position,tobin2017domain} evaluates the error feedback from the observed object in the 3D Cartesian space. This is called 3D visual servoing. These types of methods require the camera intrinsic parameters to convert all the observed object pose to a 3D coordinate system. Therefore, the accuracy of the camera calibration and the robot model will directly affect the control output. In image-based visual servoing (IBVS)~\cite{feddema1989vision,gao2015hierarchical,zheng2016image}  control, the robot will minimize the error directly in the image space. This is also called 2D visual servoing. Because of the estimation error in image space, these types methods are insensitive to the calibration errors of the cameras on the robot. The disadvantages of IBVS are lost feature points during the rotation of the robot. The singularity of the Jacobian matrix could also cause control failures. Combining the advantages of both, \citet{malis19992} proposed a 2.5D visual servoing using the homography between planar targets. Our method is related to but different from all these methods. Similar to~\citet{lee2017learning}, \textit{we use self-supervised deep learning to automatically discover the interactive matrix, thereby removing all manual calibration processes in the system}.


\section{System Design}

\begin{figure*}[!t]
    \centering
    \includegraphics[width=0.98\textwidth]{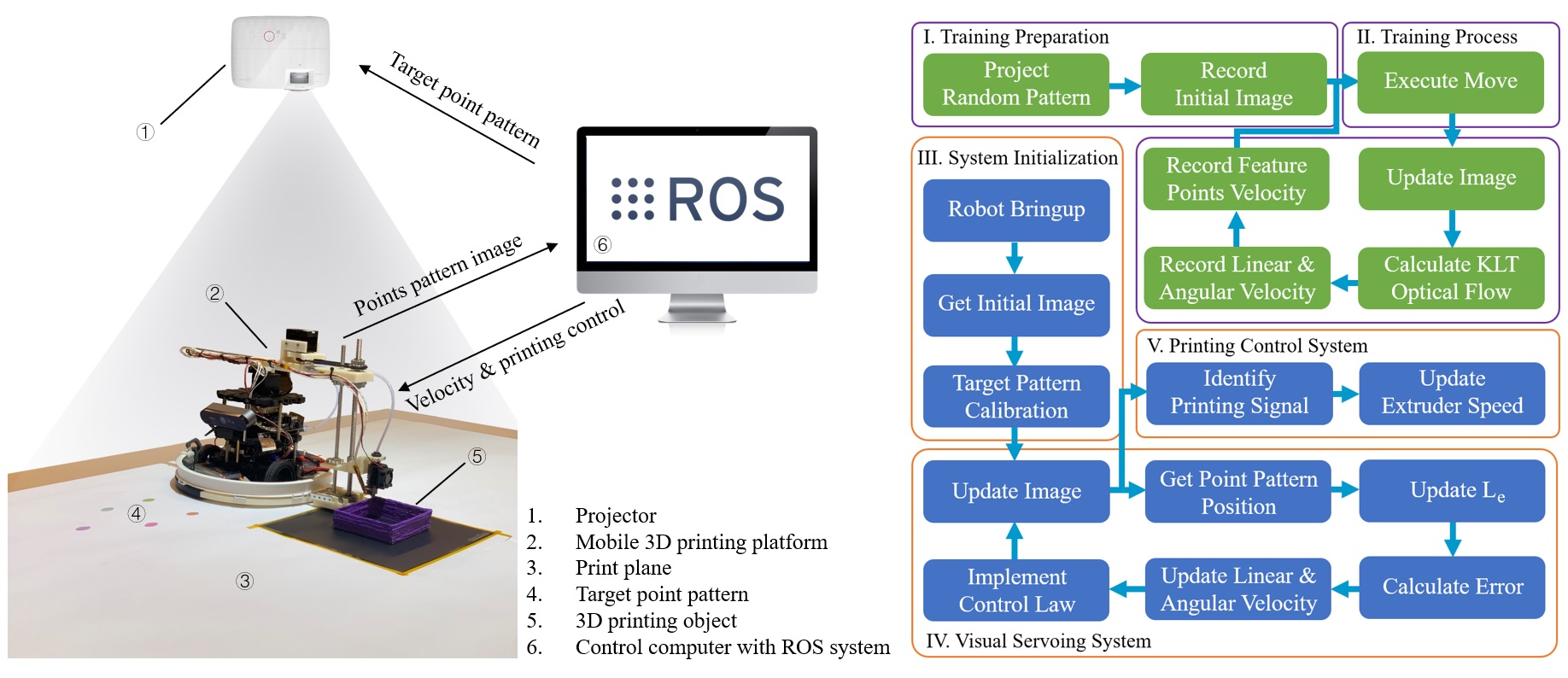}
    \caption{System settings and operation workflow. Left: projector-guided mobile 3D printing diagram. Right: training pipeline (green blocks) \&  printing pipeline (blue blocks).}
    \label{pipline}
    \vspace{-8mm}
\end{figure*}

\begin{figure}[!t]
    \vspace{3mm}
    \centering
    \includegraphics[width=0.47\textwidth]{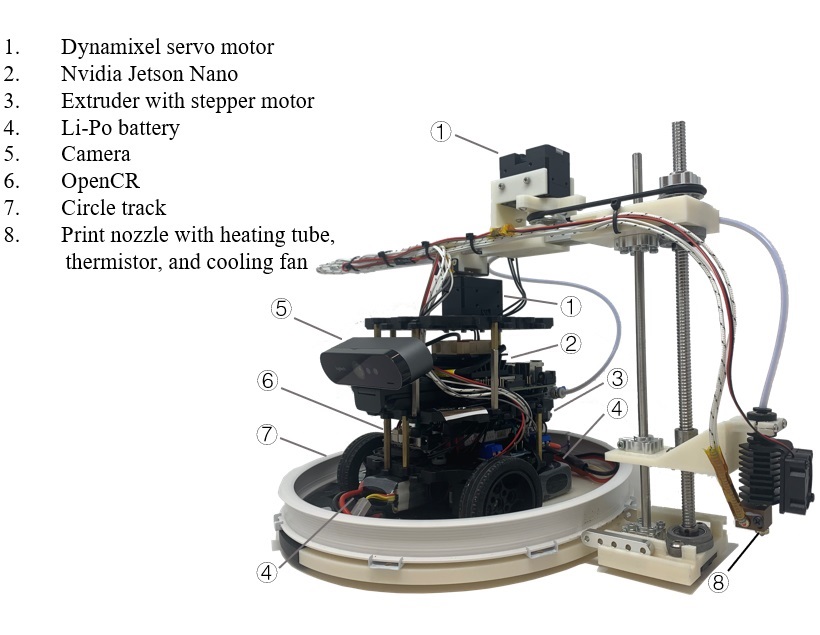}
    \caption{Hardware design of mobile 3D printing platform.}
    \label{mobile_platform}
    \vspace{-3.3mm}
\end{figure}

As shown in Figure \ref{pipline}, our mobile 3D printing system included a mobile platform, a top-down projector and a flat printing surface. 
Our mobile platform was composed of a TurtleBot and a robot arm with an FDM hotend kit. The TurtleBot provided continuous movement, and the robot arm helped the nozzle reach positions on the printing surfaces. We used an entry-level projector, BenQ MS535A SVGA, to project dynamic point pattern as the observation reference on the printing surface. The mobile platform's on-board computer computed the printing trajectory and velocity information from a projected image. The flat surface as a projection screen provided a borderless build plate for the entire mobile 3D printing system.

\subsection{Mechanical Design}

Our mobile platform was developed on Robotis TurtleBot3 burger. TurtleBot3 burger is a compact size differential drive mobile robot. It operates two dynamic wheels powered by Dynamixel XL430-W250. The original design of TurtleBot3 included stacked motors with a battery, driver board, and single-board computer(SBC) separated by four plastic plates. As shown in Figure \ref{mobile_platform}, a $360^\circ$ rotating robot arm was installed on the top layer. We expanded the bottom layer with an additional four parts parts such that the whole bottom became a circular shape. It is noteworthy that the bottom plates were composed of two symmetric semicircular 3D printing parts in our original design. However, due to the size limitations of the 3D printer, we finally divide it into four parts to meet the size limitation requirement. To avoid robot arm motion in the radial direction, a flexible curtain track was installed on the bottom plate to ensure the whole robotic arm rotated simultaneously.

\begin{figure}[!t]
    \vspace{2mm}
    \centering
    \includegraphics[width=0.46\textwidth]{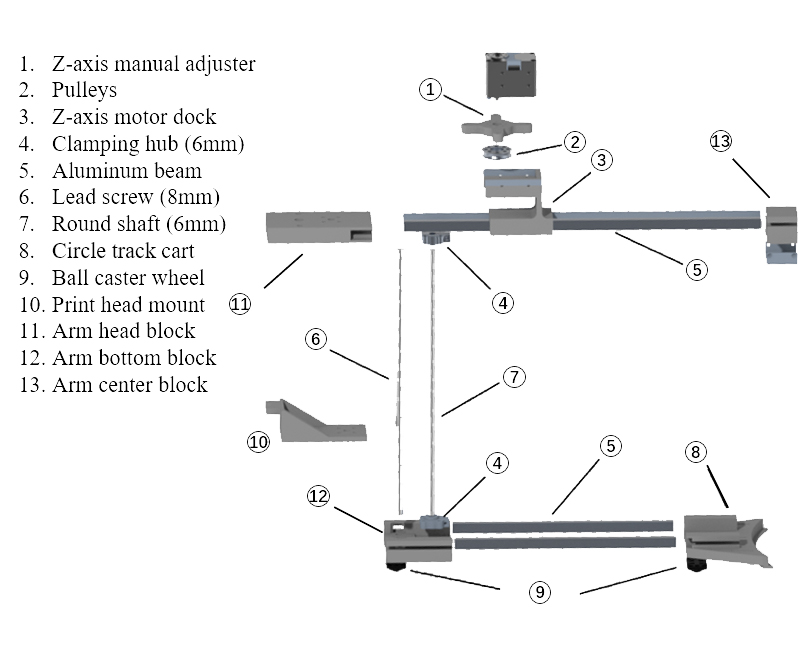}
    \caption{Adjustable length robotic arm assembly exploded view.}
    \label{robot_arm}
    \vspace{-3mm}
\end{figure}
 
The robot arm assembly consisted of two actuated DoFs. The proximal DoF is a revolute joint whose rotation axis is perpendicular to the top surface. The distal DoF is a prismatic joint that is orthogonal to the proximal DoF. Each joint is driven by a Robotis Dynamixel XM430-W350 smart servo motor. Each servo motor integrated a motor controller, network communication and wheel encoder. The robot arm's main structure has three separate ABS 3D printed parts, linked by aluminum hole pattern beams at the top and bottom. An optional circle track cart helps reduce the horizontal vibration of the robot arm when the arm length is extended in the radial direction. The repeating 3.5-mm holes on the aluminum beam provide a flexible arm length in the radial direction. The structure of the vertical prismatic joint is designed with one 8-mm lead screw and one 6-mm round shafting support rail fixed by a pillow block flange bearing. The printer head holder achieved vertical movement by connecting with the lead screw nut. A Hotend kit (Lerdge BP6 with 8-mm nozzle) is installed on the end of the printer head holder. To reduce sliding friction and better support the weight of the print head, an additional ball caster wheel is added to the bottom of the arm block.

\subsection{Mechatronic and Software System}  

Our control system is modified from TurtleBot3's mechatronic system architecture. As we previously mentioned, our main control system consisted of OpenCR (driver board) and Jetson Nano (SBC), as shown in Figure \ref{mobile_platform}. As a driver board, OpenCR is connected to four servo motors through serial communication, two for the wheels and the other two for the robot arm.
OpenCR's GPIO pins are used to control the heating tube, cooling fan, and to send the signal output to the extruder's stepper motor. Due to the different operating voltage and current settings of the actuators, we controlled the heating tube and the cooling fan through additional MOSFET breakout boards. A Thermistor (100KNTC3950) provided nozzle temperature feedback by connecting with a 4.7K Ohms pull-up resistor. Due to multiple microstepper settings, we use the TB6600 as a stepper motor driver. Two 11.1V Li-Po batteries are attached on the bottom layer, one for control circuits and the other for all actuators.

For the control software, Robot Operating System (ROS) is used to communicate with and synchronize the nodes and to handle all low-level device control. We created three additional nodes and customized messages, which worked with the original TurtleBot3 libraries. Two of the nodes are used to maintain the heating temperature of hotend and publishing the direction and speed topic for extruder control. The other node calculates the pixel errors between the desired and the captured pattern positions and sends the errors back to the main control program.
On the right of Figure \ref{pipline}, we present our training and printing workflow.


\section{Visual Servoing}

\begin{figure*}[!t]
    \centering
    \includegraphics[width=0.95\textwidth]{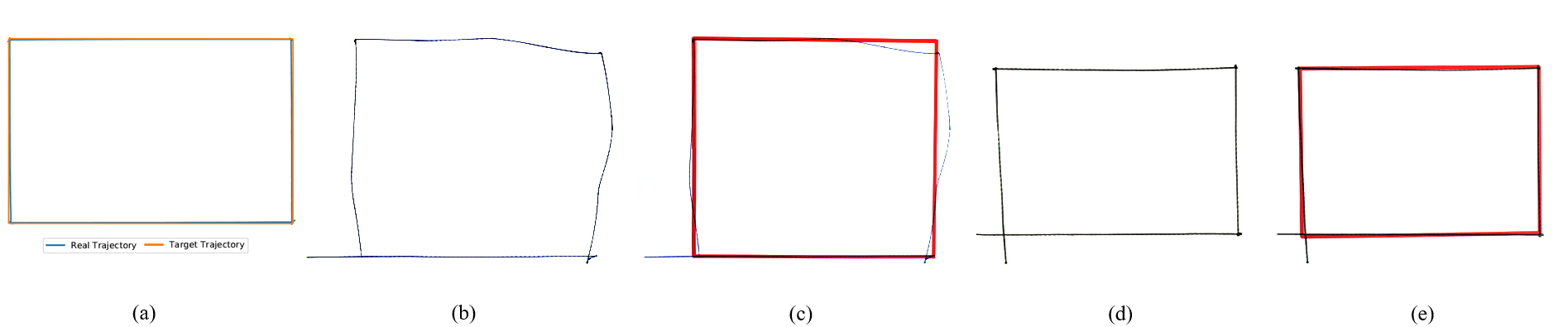}
    \caption{The trajectory following experiment results: (a) Simulation results; (b) (c) Real-world results of LBVS without control constraints; (d) (e) Real-world results of LBVS control with control constraints.}
    \label{Task_1}
    \vspace{-3mm}
\end{figure*}

\subsection{Learning-Based Visual Servoing (LBVS)}\label{sec:lbvs}

To use visual servoing to control the 2D movement of our mobile base, we need to determine the interaction matrix $\bm{L}_e \in \mathbb{R}^{2\times 2}$ between the control input $\bm{v}_\theta=[v,\omega]^T$ (i.e., linear and angular velocities) and the image pixel measurements $\bm{u}=[u_x, u_y]^T$ of a target feature point such that $\bm{\dot{u}}=\bm{L}_e \bm{v}_\theta$.
To make our mobile robot system easy to use and automatically deployable, unlike the classic IBVS or 2.5D-VS methods~\cite{chaumette2006visual,malis19992}, \textit{we want to avoid any intrinsic/extrinsic/hand-eye calibration of the camera, and any feature point depth estimation}. We achieved this using a machine-learning-base approach to find a dynamics model, as described below. Thus, we call our method LBVS.

Utilizing the prior knowledge that a homography exists between the ground plane and the camera plane, we model the interactive matrix as a function of the pixel location, i.e., $\bm{L}_e(\bm{u}): \mathbb{R}^2 \to \mathbb{R}^{2\times 2}$. If this matrix function is given, the control law of the mobile base is the same as in the classic IBVS: $\bm{v}_\theta = -\lambda \bm{L}_e^{+} \bm{e}$, where $\bm{L}_e^{+}=(\bm{L}_e^T\bm{L}_e)^{-1}\bm{L}_e^T$, $\bm{e}=\bm{u-u^*}$, and $\bm{u}^*$ is the desired image location of a feature point. Note that in a dynamic trajectory following case, $\bm{e}$ is the optical flow of the feature point from the current to the previous frame. Given $M$ tracked feature points, $\bm{e}=[\bm{e}_1^T,\cdots,\bm{e}_M^T]^T \in \mathbb{R}^{2M\times 1}$ is the stacked error vector (i.e., the optical flow), and $\bm{L}_e=[\bm{L}_e(\bm{u}_1)^T,\cdots,\bm{L}_e(\bm{u}_M)^T]^T \in \mathbb{R}^{2M\times 2}$ is the stacked image Jacobian evaluated at each feature point, and the resulting control $\bm{v}_\theta$ jointly regulates each feature point's error vector via the least squares principle.

The interaction matrix function $\bm{L}_e(\bm{u})$ can be modeled as a simple multi-layer perceptron (MLP), e.g., a ReLU MLP (2-64-64-64-4) with three 64-dimensional hidden layers. The challenge is determining how to estimate this MLP automatically. Fortunately, we can take advantage of our projector and the ground plane. During the automatic system calibration stage, we can project a set of $N$ random colored dots on the ground such that these points cover enough area on the image plane. We can control the mobile base with a sequence of $T$ frames of random velocity commands $[\bm{v}_\theta^1,\cdots,\bm{v}_\theta^T]$ while recording the color dots' image measurements at each frame as $[\bm{U}^0,\bm{U}^1,\cdots,\bm{U}^T]$, where $\bm{U}^t=[\bm{u}_1^t,\cdots,\bm{u}_N^t]$. Denoting the $i$-th point's optical flow vector at frame $t$ as $\bm{f}_i^t=\bm{u}_i^{t+1}-\bm{u}_i^t$, we can train the MLP by minimizing the L2 loss over this dataset as follows:
\begin{equation}
 \min_{\bm{L}_e} \frac{1}{T} \sum_t \frac{1}{N} \sum_i ||\bm{L}_e(\bm{u}_i^t)\bm{v}_\theta^t - \bm{f}_i^t||^2. \label{eq:lbvs}
\end{equation}

Our LBVS method has a desirable property in that it is generally applicable to a wide range of camera lenses (perspective or fisheye) and can be calibrated automatically.

\subsection{Additional Control Constraints}\label{sec:constraints}
By adding additional control constraints, our mobile platform can drive more smoothly and avoid speed overshoot. As presented above, the LBVS speed control equation is $\bm{v}_\theta = -\lambda \bm{L}_e^{+} \bm{e}$. We divide $\lambda$ into two parts based on the distance to the target point pattern. As the robot approaches to the pattern, we lower $\lambda$ to prevent speed overshoots. For large $e$, we adopt a larger $\lambda$ to reduce the error $e$ quickly. As shown in our experiments, LBVS sometimes can still lose track of the point pattern during rotation. Thus, we included an additional mode to optimize the rotation. In this mode, we can estimate $\bm{L}_e$ as long as we can obtain position information of any three points from the point pattern. We also constrain the maximum linear velocity to 0.025 m/s and maximum angular velocity to 0.05 rad/s.

\begin{figure}[!t]
    \vspace{-3mm}
    \centering
    \includegraphics[width=0.42\textwidth]{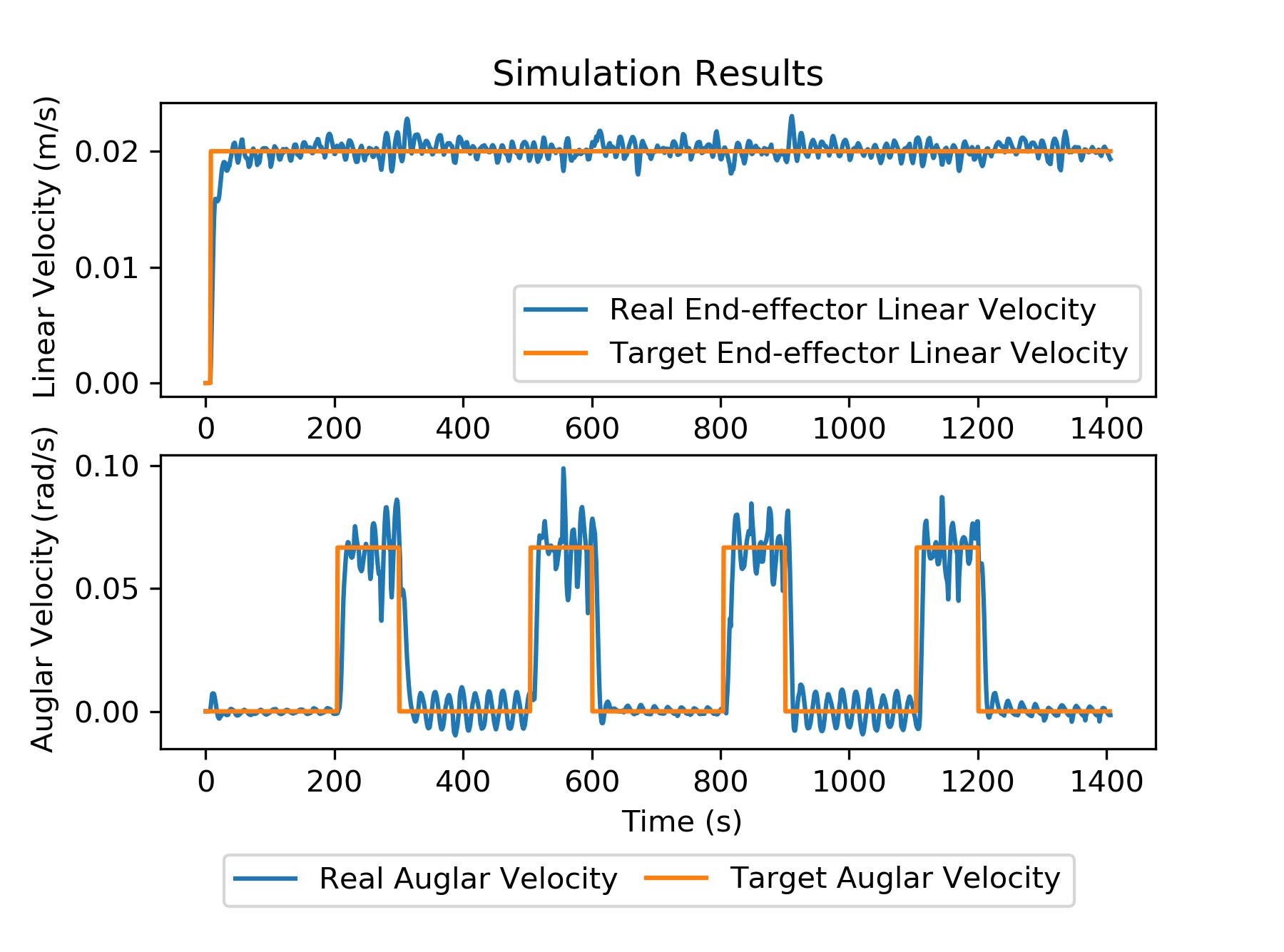}
    \caption{Simulation results of real velocity vs. target velocity.}
    \label{simulation}
    \vspace{-4mm}
\end{figure}

\begin{figure*}[!t]
    \centering
    \includegraphics[width=1\textwidth]{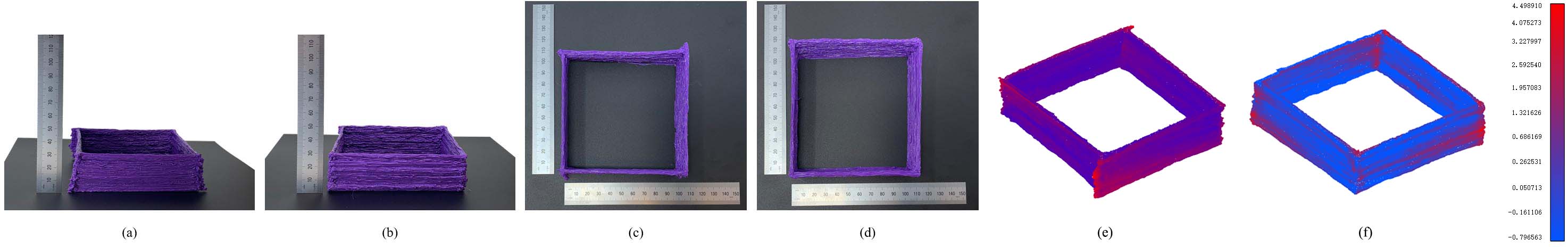}
    \caption{Printing results of single robot printing: (a),(c),(e) fixed-arm printing; (b),(d),(f) rotation compensation printing; (e),(f) comparison of scanned point cloud and ground truth.}
    \label{Task_2}
    \vspace{-1mm}
\end{figure*}

\begin{figure}[!t]
    \centering
    \includegraphics[width=0.48\textwidth]{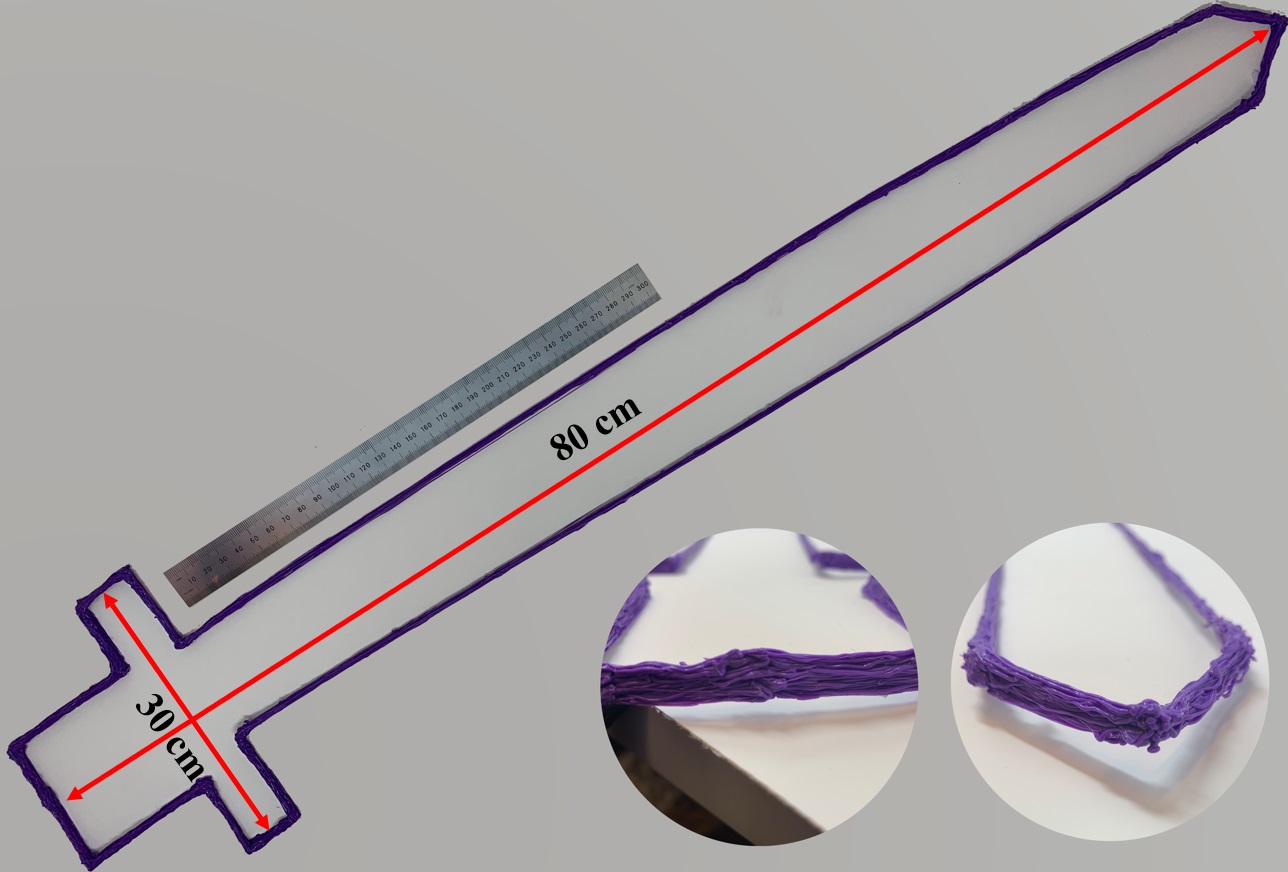}
    \caption{Large-scale collaborative printing results. Bottom circle images show the connection points of two trajectories.}
    \label{Task_3}
    \vspace{-4mm}
\end{figure}


\section{Evaluation}

We design three experiments to verify the feasibility of our mobile 3D printing system. We first introduce our experimental environment and hardware setup. In the first experiment, we test our LBVS control in the simulation and real-world environment. We use our printing system to print a hand-sized model, which could also be printed by a traditional desktop 3D printer. The purpose of the second experiment is to explore and optimize our printing system by comparing with basic 3D printing models. We verified the large-scale collaborative 3D printing in the final experiment. For each experiment, we discuss all printing deviations and failure cases that occurred.

\subsection{Experiment Setting}
\textbf{Simulation environment.}
In the simulation environment, our mobile platform will drove on a plane object on the ground. This object was used to simulate the pattern's projection by updating the surface image with moving points. We subscribe odometry topic and joint space topic for recording the position and speed of the mobile platform. We used Moveit to plan the trajectory and send the execution data to the robotic arm.

\textbf{Real-World Environment.}
We mounted a projector on the ceiling and adjusted the keystone correction. To ensure that the printing surface was the flat and level, we used three $0.61\ m\times 1.22\ m$ medium-density fiberboard (MDF) to form a $1.82\ m \times 2.44\ m$ flat surface. We laid roller paper on the MDF to replace the yellow wood color. The white background allowed the camera to capture the feature points more easily. For easy peeling off of the printing model, we set the Buildtak printing board on the paper surface in single robot printing experiment.

\textbf{Data Collection and Network Training.}
We used the well-known KLT algorithm~\cite{lucas1981iterative} to calculate the optical flow $[\dot{u_x}, \dot{u_y}]^T$ between image frames. In our real world experiment, we projected 10,000 little triangles on the printing surface with a uniform distribution. We set a minimum requirement of 20 feature points that need to be capture. The real linear and angular velocity $[v, \omega]^T$ can be read directly from the joint space ROS topic. As explained in Section~\ref{sec:lbvs}, we let the robot move straight forward and backward with no angular velocity while recording $v$ and $[\dot{u_x}, \dot{u_y}]^T$ for different image points. We then let the robot spin without linear velocity while recording $\omega$ and $[\dot{u_x}, \dot{u_y}]^T$ for different image points. Finally, we combined the two datasets to train the model prediction $\bm{L}_{e}$ via equation~\eqref{eq:lbvs}. In the training process, we used $10^5$ feature points and $3000$ frames, as $T$ in equation~\eqref{eq:lbvs}. The data collection and network training were completed in 30 minutes with $1\bm{e}^{-4}$ learning rate by using an Nvidia TitanV graphics card.

\subsection{Trajectory and Control Accuracy}

In this experiment, we designed a rectangular motion trajectory. In Gazebo, the results showed that the LBVS could control the mobile platform to complete the trajectory well. In Figure \ref{simulation}, both the linear and angular velocities of the mobile platform exhibited almost the same shapes as the pattern's speed. Meanwhile, as expected, there was a small delay between them. The reason was that the mobile platform needed to observe and process the point pattern before reacting.

For the real-world environment, we first tested the LBVS without additional control constraints. At the beginning of the trajectory, the mobile platform presented the same performance as that in simulation. After the pattern rotated, there were several moments when the robot maintained a constant angular velocity due to lost points. Therefore, deviations occurred in the upper right corner, as shown in Figure \ref{Task_1} (b). Furthermore, we found that the mobile platform moved back during its rotation process. These two factors could strongly impact the printing quality and cause printing failure during the  process.

To optimize the LBVS control, we added the control constraints described in Section~\ref{sec:constraints}. As shown in Figure \ref{Task_1} (d) and (e), we only achieved a slight adjustment at the top edge. From Table~\ref{tab:label_real_ate}, the absolute trajectory error (ATE)~\cite{zhang2018tutorial} also indicated that the error of the LBVS with constraints was much smaller than that of the LBVS without constraints.

\begin{table}[!t]
    \vspace{1mm}
    \centering
    \caption{ATE for real-world experiment}
    \label{tab:label_real_ate}
    \resizebox{0.48\textwidth}{!}{%
    \begin{tabular}{c| c c c }
    \toprule
    ATE (\si{mm}) & RMS & Median & Max \\ \midrule
     Without Constraints & 19.28 & 10.13 & 57.67\\
     With Constraints & \textbf{1.67} & \textbf{3.12} & \textbf{7.30}\\
    \bottomrule
    \end{tabular}
    }
    \vspace{-3mm}
\end{table}

\subsection{Single Robot Printing}
We used a single mobile platform to print a cuboid model. Based on the previous experiments, we learned that the rectangle corners were the most critical positions affecting the printing results. To better handle four corners, we designed two different corner printing methods. The first method fixed the proximal joint position to $-\pi/2$ and completely relied on extruder control. The other method used the robotic arm to compensate for the rotation of the robot chassis. It kept the end effector stationary when the mobile platform turned. Additionally, we propose a new printing method that could optimize our printing results. In this method, the mobile platform prints front and back three times on every edge once before it turns to the next edge. The models shown in Figure \ref{Task_2} were not completed all at once. Based on the previous printing results, we adjusted the extruder control and replaced the batteries for every 1-cm height. We measured the printed wall thickness in Table~\ref{tab:label_real} in comparison with the designed thickness as ground truth (GT).

As Figure \ref{Task_2} shows, the model printed by the fixed robot arm method significantly over-printed at the four corners. When the mobile platform completed the turn, the end effector could not always reach the previous printing endpoint. We found that the model surface printed by the rotation compensation method was cleaner and smoother due to less over-stacking at the model's four corners.

\subsection{Large-Scale Collaborative Printing}

The purpose of the final experiment was to test our mobile 3D printing system, which could quickly set up multiple printing platforms and complete large-scale printing. Here, we used two mobile platforms to complete this experiment. We designed an asymmetric contour, a sword, which is shown in Figure \ref{Task_3}. This sword model had a total length of 0.80 m and a width of 0.30 m, and multiple corners that need to be turned. Each mobile platform needed to print half of the entire shape at the same time. The printing model had stacked layers at the sword's point and grip, as shown in the Figure \ref{Task_3} circle. Furthermore, we printed directly on the laid roller paper instead of the traditional 3D printing surface.

During the experiment, we found that most of the mobile printing failures also occured with the desktop 3D printer. The most common failure is model shrinkage. Our model separated from the printing surface several times. To overcome this problem, we overprinted at every corner except the point of the sword. Avoiding collisions and control interference are two other issues that require concern. In this experiment, we manually designed the printing trajectory to avoid the collision. We also change the color of the point pattern to avoid mutual control interference.

\begin{table}[!t]
    \centering
    \caption{Wall thickness for real-world experiment}
    \label{tab:label_real}
    \resizebox{0.48\textwidth}{!}{%
    \begin{tabular}{c| c c c c c }
    \toprule
     Wall Thickness (\si{mm}) & GT & Mean & Maximum & Minimum & St.Dev \\ \midrule
     Without Constraints &  \multirow{2}*{2.00} & 3.72 & 4.92 & \textbf{2.33} & 0.40\\
     With Constraints & & \textbf{3.02} & \textbf{4.17} & 2.74 & \textbf{0.24}\\
    \bottomrule
    \end{tabular}
    }
    \vspace{-3mm}
\end{table}


\section{Conclusions}

In this paper, we proposed a projector-guided non-holonomic mobile 3D printing system. Compared with a traditional 3D printer, we overcame the 3D printing size limitation due to the build plate and gantry structure in conventional 3D printing. Compared with prior works, our method does not require any manual calibration or any world coordinate for the mobile robot's pose feedback. From the experimental results, our LBVS worked effectively on the non-holonomic mobile platform.

\textbf{Limitations and Discussions.}
Although our mobile 3D printing system has a great potential for future 3D printing and robotic construction, our current system prototype is still far from producing a high quality model with satisfactory surface smoothness, accurate size dimension and shape diversity. Our deviation analysis showed that the material overfill and layer shifting occurred on almost every layer. Currently, we only printed a 3D model’s outer surface. When more degrees of freedom are added to the robot arm, our system should be able to print more diverse shapes with even internal structures. Also our current projector's illumination level is a limitation if we want to use this system in an outdoor environment, where laser projectors would be more helpful.

\textbf{Future Work.}
In the future, we will optimize the LBVS control for more DOFs, update our hardware design, and print 3D models with competing qualities to those of desktop 3D printers. We believe that our system could also use different materials and operate in different application scenarios ranging from desktop-scale 3D FDM printing to room-scale 3D construction printing with multiple projectors.


\section*{Acknowledgments}

This research is supported by NSF CPS program under CMMI-1932187.


\addtolength{\textheight}{-0.2cm}   


\bibliographystyle{IEEEtranN}
\bibliography{root}

\end{document}